\pdfoutput=1

\documentclass[11pt]{article}

\usepackage[]{EACL2023}

\usepackage{times}
\usepackage{latexsym}
\usepackage{amsmath}
\usepackage{amssymb}
\usepackage{amsfonts}
\usepackage{latexsym}

\usepackage{booktabs}
\usepackage{multirow}
\usepackage{graphicx}
\usepackage{url}

\usepackage[T1]{fontenc}

\usepackage[utf8]{inputenc}

\usepackage{microtype}

\usepackage{inconsolata}

\title{Exploring the Limits of Transfer Learning with Unified model in the Cybersecurity Domain}

\author{Kuntal Kumar Pal \\ kkpal@asu.edu \And
        Kazuaki Kashihara \\\texttt{kkashiha@asu.edu}\And
        Ujjwala Anantheswaran \\ \texttt{uananthe@asu.edu}
        \AND
        Kirby C. Kuznia \\ \texttt{kkuznia@asu.edu} \And
        Siddhesh Jagtap \\\texttt{sjagtap5@asu.edu} \And
        Chitta Baral \\ \texttt{chitta@asu.edu} \\
        \AND
        School of Computing and Augmented Intelligence, Arizona State University}

\begin{document}
\maketitle
\begin{abstract}
With the increase in cybersecurity vulnerabilities of software systems, the ways to exploit them are also increasing. Besides these, malware threats, irregular network interactions and discussions about exploits in public forums are also on the rise. To identify these threats faster, to detect potentially relevant entities from any texts and to be aware of software vulnerabilities,  automated approaches are necessary. Application of natural language processing (NLP) techniques in the Cybersecurity domain can help in achieving this. However, there are challenges such as, the diverse nature of texts involved in the cybersecurity domain, unavailability of large-scale publicly available datasets, and significant cost of hiring subject matter experts for annotations. One of the solutions is building multi-task models that can be trained jointly with limited data. In this work, we introduce a generative multi-task model, Unified Text-to-Text Cybersecurity (UTS), trained on malware reports, phishing site URLs, programming code constructs, social media data, blogs, news articles and public forum posts. We show UTS improves performance of some cybersecurity datasets. We also show that with few examples UTS can be adapted to novel unseen tasks and nature of data.  
\end{abstract}

\section{Introduction}

In recent times, increasing cybersecurity risks, malware threats, and ransomware attacks are getting more dangerous and common. There are often discussions about them in public forums~\cite{istr/LiCHCN21,isi/AlmukayniziMNSS18,cyconus/AlmukayniziNDSS17} and social media~\cite{sbp-brims/ShuSS018, ccs/HuangW20} both before and after attacks. Private companies also release detailed malware reports such as Symantec~\cite{black-vine2015} and Cylance~\cite{op-dust-storm2016}. In addition, government agencies (NIST) and other non-profit organizations keep reports of software vulnerabilities  through NVD\footnote{\url{https://nvd.nist.gov/}} and MITRE\footnote{\url{https://cve.mitre.org/}} respectively to prevent exploitations. 
Natural language processing (NLP) methods can help in reducing the potential threats by identifying texts mentioning cybersecurity vulnerabilities or malicious exploits (\textit{text-classification}), extracting mentions of relevant entities of threats in public discussions (\textit{named entity recognition}) or identifying the relation of threats with other entities (\textit{relation classification}). It can be used to extract a threat or an event from a text (\textit{event detection}) along with its arguments (\textit{event argument extraction}) to find its source or estimate its damage.

In recent years, natural language (NL) domains have seen considerable improvements in all the natural language understanding (NLU) tasks with many powerful transformer-based models such as BERT~\cite{devlin2018bert}, RoBERTa~\cite{liu2019roberta}, and XLNet~\cite{NIPS/YangDYCSL19}. With improvements in the NL domain, these models have been adapted and shown to improve performance in other domains such as BioBERT~\cite{BioBERT/bioinformatics/LeeYKKKSK20}, mimicBERT~\cite{MimicBERTjournals/corr/abs-2003-07507}, ClinicalBERT~\cite{clinicaBERT/journals/corr/abs-1904-05342}, blueBERT~\cite{peng2019transfer} in the biomedical domain, sciBERT~\cite{beltagy2019scibert} in the scientific domain (computer science and biomedical), LegalBERT~\cite{chalkidis2020legal} in the legal domain and FinBERT~\cite{ijcai/0001HH0Z20} in the financial service domain. Motivated by these approaches, we introduce a unified model in the cybersecurity domain capable of performing multiple NL tasks.

Unlike other domains, in Cybersecurity domain the nature of texts is quite diverse (natural language text, URLs, malware reports, system calls, source code, binaries, decompiled code, network traffic, software logs~\cite{phandi-etal-2018-semeval,kirillov2011malware,Queiroz2019DetectingHT,Marchal2014PhishStormDP,aaai/SatyapanichFF20-CASIE,NER-BridgesJIG13,uss/ChuaSSL17,sp/ZhangYYTLKAZ21,sigsoft/PeiGBCYWUYRJ21}). This led to the introduction of specific models capable of performing individual tasks like cyber-bullying detection CyberBERT~\cite{McDonnell2021CyberBERTAD} and cybersecurity claim classification CyBERT~\cite{Ameri2021CyBERTCC}. Apart from this, there is a scarcity of large-scale publicly available annotated datasets. These challenges demand the need of developing \textit{robust} models capable of performing \textit{multiple tasks} by \textit{learning from many datasets together}. Hence, we introduce an \textbf{U}nified, \textbf{T}ext-to-Text Cyber\textbf{S}ecurity ($UTS$) model.

In this work, a transformer-based generative model, T5~\cite{DBLP:journals/corr/abs-2005-14165}, is trained in a multi-task setting on \textit{eight} fine-grained NLP tasks involving \textit{13} datasets in the cybersecurity domain. We used task based prompt prefixes to help the model to learn the task instead of learning specific datasets. Our goals is to make the model more robust by training on a variety of texts. We show the model's generalizability on unseen tasks (task transfer) and on unseen datasets (domain transfer) in three few-shot settings. 
In the spirit of open science, we will release our research artifacts, including all processed datasets, the source code, and our trained models, upon  acceptance.

We summarize our contributions as follows. We 
\begin{itemize}
    \item Propose a unified text-to-text transformer model (UTS) in the cybersecurity domain which is capable of performing four fundamental NLP tasks and their sub-tasks. To the best of our knowledge, this is the first attempt to unify varied text nature in this domain. 
    \item Establish a benchmark of \textit{13} existing cybersecurity datasets processed in text-to-text format involving \textit{eight} NLP tasks for future models to compare with.
    \item Perform extensive experiments with UTS to assess its ability to adapt to novel task and nature of texts in three few-shot settings.
    
\end{itemize}









\begin{table*}[ht!]
\resizebox{\textwidth}{!}{%
\begin{tabular}{@{}lllllrr@{}}
\toprule

\multicolumn{1}{c}{\textbf{Dataset}} & \multicolumn{1}{c}{\textbf{Nature}} & \multicolumn{1}{c}{\textbf{Cybersecurity Task}} & \multicolumn{1}{c}{\textbf{Mapped NLP Task}} & \multicolumn{1}{c}{\textbf{Dataset Identifier}} & \multicolumn{1}{c}{\textbf{\#Samples}}  & \multicolumn{1}{c}{\textbf{\#Class }} \\ \midrule
MalwareTextDB-$V_2$ \cite{phandi-etal-2018-semeval} &
  APT Reports &
  Malware Text Detection &
  Sentence Classification &
  MDB-SENTCLS &
  12,736 &
  2 \\
  
MalwareTextDB-$V_2$ \cite{phandi-etal-2018-semeval} &
  APT Reports & 
  Malware Entity Relation Identification &
  Relation Classification &
  MDB-RELCLS &
  10,802 & 
  4\\

  CyberThreatDetection \cite{Queiroz2019DetectingHT} &
  Public Forum Posts &
  Hacker's Threat Detection &
Text Classification &
  CTD &
  12,575 &
  2 \\
SMS Spam \cite{doceng/AlmeidaHY11} &
  Text Messages &
  Spam Message Detection &
  Text Classification &
  SMS-SPAM &
  5,574 &
  2 \\

Phishstorm \cite{Marchal2014PhishStormDP} &
  URLs & 
  Phishing URL Detection &
  Text Classification &
  URL &
   95,911 &
  2  \\
  
  Soft-Flaw CLS \cite{DVN/1TCFII_2020}) 
& Social Media (Twitter) 
& Vulnerable Tweet Detection
& Text Classification 
& Soft-Flaw CLS            
   
&  1,000
& 2 \\

  CASIE \cite{aaai/SatyapanichFF20-CASIE} &
  CS News Articles &
Event Argument Role Identification  &
  Token Classification &
  CASIE-ARGROLE &
  11,222 &
  13 \\  \hline
  
Stucco Auto-labelled \cite{NER-BridgesJIG13} &
  NVD-CVE Descriptions &
  Information Security Entities Extraction &
  Named Entity Recognition &
  SAL &
  15,192 &
  15 \\
  
  Soft-Flaw NER \cite{DVN/1TCFII_2020}
& Social Media (Twitter) 
& Cybersecurity entity Detection   
& Named Entity Recognition 
& Soft-Flaw NER     
&  826
& 1 \\

SOFTNER \cite{Tabassum20acl} &
 Text with Source Codes &
 Computer Programming Entity Extraction &
 Named Entity Recognition &
 SOFTNER &
  24,092 &
  20
  \\ 
  
   \hline
  CASIE \cite{aaai/SatyapanichFF20-CASIE} &
  CS News Articles &
Event nuggets(keywords) Extraction &
  Event Extraction &
  CASIE-EVTDET &
  16,230 &
  5 \\
  
  CASIE \cite{aaai/SatyapanichFF20-CASIE} &
  CS News Articles &
Detect arguments of event from sentence &
  Event Argument Extraction &
  CASIE-ARGDET &
  17,956 &
  21 \\\hline
  
  CVSS \cite{9680155} &
  CVE Description &
Vulnerability Impact Score Estimation &
  Regression &
  CVE-IMPACT & 
  48,827 & -
   \\
   
 \bottomrule
\end{tabular}%
}
\caption{Dataset Descriptions with eight fine-grained NLP tasks. \#Samples represents full dataset samples}
\label{tab:datasets}
\end{table*}



\section{Approach}
\label{sec:approach}
\begin{figure*}[ht!]
    \centering
    \includegraphics[width=.95\textwidth]{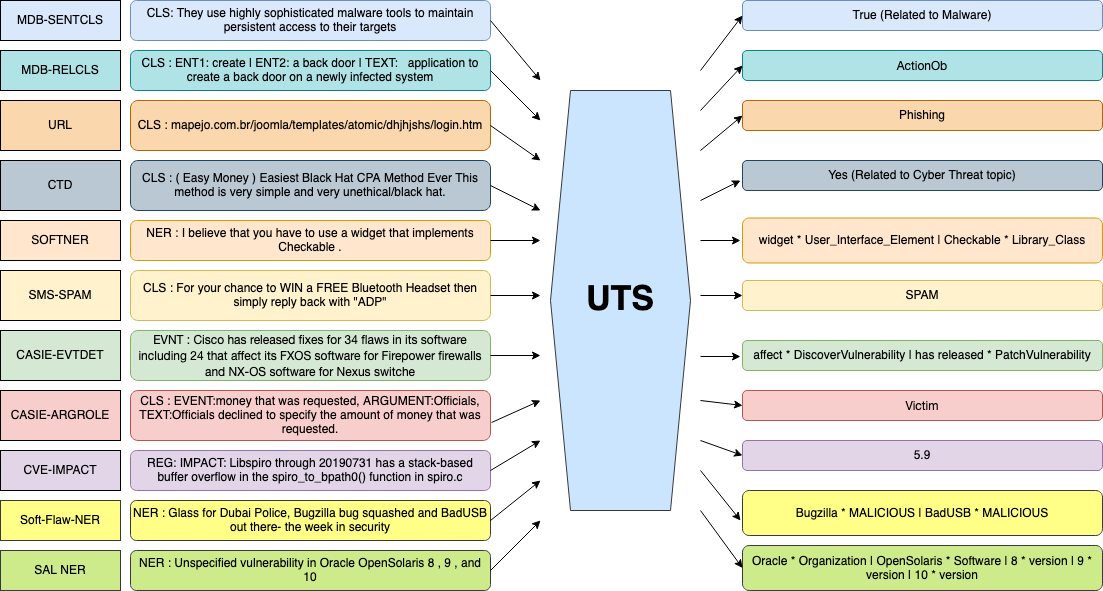}
    \caption{Illustration of $UTS$ (Unified Text-to-Text CyberSecurtiy) model}
    \label{fig:model_diagram}
\end{figure*}

We develop a generative transformer based model (UTS) trained on various nature of texts in multi-task setting to perform the fundamental NLP tasks like classification (CLS), named entity recognition (NER), event detection (ED) and regression (REG) together. We assign task-based control code (prompts) to teach the model different tasks. Our approach can be seen from Figure \ref{fig:model_diagram}.

\noindent
\textbf{Generative approach:}
We consider T5-base as the underlying model of UTS.
This generative text-to-text approach helps us to formulate various NLP tasks into a uniform input-output format and train together with multiple tasks. 
For CLS tasks, we train the model to generate the exact class names for the given input. For NER and ED tasks, the model needs to extract the entities in a given text along with their types. So, we train the models to generate a concatenation (using `|') of entity name along with its type (separated by `*'). The model is trained to generate the exact regression scores for REG task.

\noindent
\textbf{Multi-Task Training:}
All the training datasets of these four fundamental NLP tasks - CLS, NER, ED and REG - are grouped together for joint training with the hypothesis that in this way the model can learn from more examples of the same task and similar examples of multiple tasks. 
Under CLS task, there are four fine-grained classification tasks: Text, Sentence, Relation, and Token Classification. 
We parse the textual output generated by the model and evaluate $UTS$ on test data of each of the corresponding datasets. To avoid confusion of the model in identifying similar yet textually different categories, we use unique mapping of the entity types across all extraction tasks.

\noindent
\textbf{Prompt-Based Approach:} We use task-based control codes as prompt-prefix for training the models in a multi-task setting so that it learns to perform each task instead of learning from any particular dataset.
We prepend task acronyms CLS, NER, EVNT, REG with the input for classification, named entity recognition, event detection and regression tasks respectively.




\noindent
\textbf{Problem Formulation:}
The problem formulation is defined here.
Given an input text $I$ = \{$i_1$, $i_2$, ..., $i_n$\} and a task $T$, the model should generate a stream of output tokens $O = \{o_1 | o_2 | ... | o_n\}$ defined by the task. For CLS and REG tasks, $O=\{o_1\}$, which represents the class-name and floating-point value respectively. For NER and ED tasks, each $o_i$ represents entity and entity-type separated by a pre-defined marker i.e. $o_i=\{e_i * t_i\}$. The task (T) is formulated as an instruction to help the models to learn individual tasks in this setting.

\section{Dataset Preparation}
We prepare \textit{13} datasets involved in \textit{eight} NLP tasks. The summary of the datasets is presented in Table \ref{tab:datasets}. There are four fundamental tasks in the collected datasets; Classification, Event Detection, Named Entity Recognition, and Regression respectively.
For each of the datasets, we used the original test splits if mentioned in the paper. Otherwise, we consider 20\% of the data chosen randomly as the test split. The details of each dataset are as follows.

\subsection{Classification}

\textbf{MalwareTextDB-V2:} This dataset
\cite{phandi-etal-2018-semeval} is constructed from 83 APT reports. Each report  contains multiple paragraphs regarding various activities of malwares. We consider two tasks from this dataset for $UTC$. They are : (1) \textit{Sentence Classification} - classifying whether individual sentences are relevant to cybersecurity applications, and (2) \textit{Relation Classification} - classifying the relation between two given entities.
We take 68 documents as train and 15 documents test datasets. Each document has multiple sentences which we pre-process as each input sample. 


\noindent
\textbf{SMS-SPAM:}
Another classification subtask is to classify the spam messages. This benchmark dataset~\cite{doceng/AlmeidaHY11} is for detecting SMS spam messages. 
The SMS-SPAM dataset is a combination of several publicly available SMS corpora and websites. 

\noindent
\textbf{CyberThreatDetection:}
This dataset \cite{Queiroz2019DetectingHT} was constructed from various hacker forums. Posts were collected and labeled by humans into 3 categories. \textit{Yes}, for posts that appear as malicious posts. \textit{No}, for posts not related to hacker activity. \textit{Undecided}, for posts where the annotator did not have enough information. The original authors counted the \textit{Undecided} labels as \textit{Yes} labels. 

\noindent
\textbf{PhishStorm:}
This dataset \cite{Marchal2014PhishStormDP} includes around 96k URLs. These URLs are labeled as normal or phishing, and were collected through PhishTank,\footnote{\url{http://www.phishtank.com}} which is a crowd sourced project where people submit phishing URLs and were later confirmed by several people. 

\subsection{Event Detection}

\textbf{CASIE:} This is the first cybersecurity Event Detection dataset \cite{aaai/SatyapanichFF20-CASIE} with five main types of events. We consider three tasks from this dataset: (1) \textit{Event Extraction} (2) \textit{Event Argument Detection} and (3) \textit{Event Argument Role Detection}. 
Event Extraction is a task to extract event nuggets that are words or phrases that best express the event occurrence clearly. 
Event Argument Detection is a task to detect event arguments that are event participants or property values. They can be tangible entities involved in the event such as person or organization, or attributes that specify important information such as time or amount.
Event Argument Role Detection is a task to find roles between given event nuggets and event arguments. A role is a semantic relation between an event nugget and an argument. Thus, each event type specifies the roles it can have and constraints on the arguments that can fill them.


\subsection{Named Entity Recognition}

\textbf{Stucco-Autolabeled:} This dataset \cite{NER-BridgesJIG13} is constructed from Common Vulnerabilities and Exposure (CVE) databases containing descriptions of information security issues from Jan, 2010 to Mar 2013. In Stucco-Autolabeled dataset, each word in the corpus is auto-annotated with an entity type. 
This dataset has 15 entity types.



\noindent
\textbf{SOFTNER:} This dataset~\cite{Tabassum20acl} has 20 annotated entity types from 1237 StackOverflow QA pairs. The text is embedded with source code constructs from many programming languages.

\noindent
\textbf{Soft-Flaw NER:} Cybersecurity NER corpus 2019 corpus~\cite{DVN/1TCFII_2020} consists of 1000 annotated tweets. 
The entities marked are usually the name of the software/system/device/company with a security related issue, or the name of a malware. There is a corresponding classification dataset as well (Soft-Flaw CLS). 

\subsection{Regression}
\textbf{NVD CVE metrics:} The NIST National Vulnerability Dataset uses vulnerabilities found through the CVE (Common Vulnerabilities and Exposure) system. Human security experts assign a corresponding CVSS (Common Vulnerability Scoring System) vector, and from that, the exploitability and impact score for the vulnerability is calculated. We split the data from 2002 onward into train and test in a 1:1 proportion as per the previous work~\cite{9680155} and directly generate the scores from the descriptions.

We describe our Unified Model datasets and Transfer Learning datasets in the next subsection.

\subsection{Unified Model Datasets} 
\label{sec:uni_dataset}
Out of 13 datasets in Table \ref{tab:datasets}, we jointly train UTS on 10 datasets: MDB-SENTCLS, MDB-RELCLS, URL, CTD, SMS-SPAM, and CASIE-ARGROLE for classification, CASIE-EVTDET for event detection, SOFTNER and SAL for named entity recognition and CVSS-IMPACT for regression task. We consider full volume of each of these dataset for unified training. 

\subsection{Transfer Learning Datasets}

\noindent
\textbf{Task Transfer:}
We prepare Entity Extraction (EE) and Entity Typing (ET) tasks from two NER datasets; SAL, SOFTNER. We also prepare event argument extraction (EAE) and event argument typing (EAT) tasks from CASIE-ARGDET dataset. From Soft-Flaw NER dataset, we only prepare dataset for EE task since there is only one entity type `Malicious'. We do not consider Soft-Flaw during unified training since the dataset volume is small. To prepare the few-shot datasets, we randomly pick at least one sample per type from EE and EAE tasks (if the number of types is more than the size of few-shot dataset, we randomly pick a subset) to make  label-balanced data.


\noindent
\textbf{Domain Transfer:}
We only consider Soft-Flaw dataset for this experiment since the nature of texts is unique as compared to other datasets. This dataset is prepared from social media (twitter) texts and can be used to see the adaptability of UTS on different nature of texts. To prepare the few-shot datasets, we make sure that the positive and negative samples are balanced.




\section{Experiments}

\subsection{Unified Experiments}
First, we pre-process the training data of each of these 10 datasets into text-to-text format as mentioned in subsection \ref{sec:uni_dataset}. Then, we train T5-base in a multi-task setting with the prepared training data. After the training, we evaluate the trained model (UTS) on individual test datasets. In addition, we compare the performance with existing best models and T5-base trained individually with each dataset.

\subsection{Few-shot Experiments}
We consider three few-shot settings (FS-20, FS-50 and FS-100) based on the number of examples (20, 50, 100) on which UTS is trained on.

\noindent
\textbf{Task Transfer:}
We experiment to see whether UTS can adapt to novel tasks from another known task on 3 few-shot settings and compare with T5-base trained on full dataset. To understand the extent of task transfer by UTS, we consider three few-shot sub-categories: (1) \textit{Domain Known Task Related (DKTR)}: trained model has knowledge of the data and NER task but has not learnt entity extraction (EE) and entity typing (ET) tasks (2) \textit{Domain Known Task Unrelated (DKTU)}: trained model knows the data and how to perform event detection (ED) task and event argument role classification task but does not know event argument extraction (EAE) and event argument typing (EAT) tasks. Here, Argument Role and Argument Types classes are different and (3) \textit{Domain Unknown Task Related (DUTR)} - trained model knows NER task but neither knows EE task nor has seen the data.

\noindent
\textbf{Domain Transfer:}
We experiment whether UTS can adapt to a different textual nature input for a known task. We consider social media (twitter) dataset, Soft-Flaw (both CLS and NER tasks), for this experiment in three few-shot settings where we train UTS with 20, 50, and 100 training samples and evaluate on full test data. In addition, we compare UTS with T5 trained on full training data. For the Soft-Flaw CLS dataset, zero-shot experiment is done to see if UTS can adapt to a different text nature without training on any examples.

\subsection{Metrics}
The generated output string is parsed and evaluated based on the task. For all variations of classification and regression, we consider an \textit{exact match} between the generated and original gold output. For extraction tasks, we parse generated outputs based on the predefined markers to get the entity sets (entity name and entity type). We report weighted F1 scores for all the tasks except the regression tasks where we consider exact-match accuracy as the metric for evaluation.

\subsection{Experimental Setup}
We use T5-base (220M parameters) for $UTS$. We set a predefined training budget of 30 epochs and hyperparameter tuning for our experiments. We train with 5e-5 learning rate, and 0.01 warm-up ratio. We perform the experiments with four 81GB Nvidia A100 GPUs with training batch size of 12. We consider beam size of 4. The average training time is $\sim$24hrs.


\begin{table}[]
\small
\centering
\begin{tabular}{@{}lccc@{}}
\toprule
Dataset       & Previous Best & T5 & UTS        \\ \midrule
MDB-SENTCLS   & 57.00$\lozenge$   & 84.04   & 84.44 \\
MDB-RELCLS    & 85.70$\lozenge$ & 99.79   & 99.69 \\
CTD$\ddag$           & 93.00$\blacklozenge$   & 92.17   & 92.00             \\
SMS-SPAM$\dagger$      & 91.90$\bigstar$ & 99.45   & 98.54          \\
URL$\dagger$           & 94.70$\clubsuit$ & 98.99   & 99.01 \\
SAL           & 93.40$\heartsuit$ & 98.46   & 97.60           \\
SOFTNER       & 79.10$\triangle$ & 72.90    & 77.02          \\
CASIE-EVTDET  & 79.90$\spadesuit$ & 81.43   &  83.53 \\
CASIE-ARGROLE$\dagger$ & 82.90$\spadesuit$ & 91.67   & 92.50  \\
CVE-IMPACT    & NA   & 76.58   & 76.95 \\ \bottomrule
\end{tabular}
\caption{Performance (wtd F1 score) of UTS  compared to T5-base trained on individual datasets and previous best. 
$\ddag$ represents the performance is compared with Positive Recall. 
$\dagger$ represents the performance is compared with macro-F1 score while weighted-F1 score for the rest. 
The Previous Best scores are from the following works respectively; $\lozenge$~\cite{phandi-etal-2018-semeval},$\blacklozenge$~\cite{Queiroz2019DetectingHT}, $\bigstar$~\cite{webist/MohassebAK20}, $\clubsuit$~\cite{Marchal2014PhishStormDP}, $\spadesuit$~\cite{aaai/SatyapanichFF20-CASIE}, $\heartsuit$~\cite{simran2019deep}, and $\triangle$~\cite{Tabassum20acl}.}
\label{tab:sota}
\end{table}


\section{Results and Discussion}

In this work, we seek to answer the following research questions through various experiments.

\smallskip
\smallskip
\noindent
\textbf{R1: How does UTS perform, compared to T5 and previous best?}
Table \ref{tab:sota} shows the performance of T5 trained on individual training data compared to UTS trained on all 10 datasets in a multi-task setting. We find that SOFTNER and CASIE-EVTDET shows 4\% and 2\% improvements respectively. For rest of the tasks, the performance change is marginal and most importantly it does not drop significantly. Thus, the trained UTS model has the understanding of four fundamental NLP tasks, and has seen multiple nature of texts as well.

In addition, we show how UTS performs as compared to previous best approaches in Table \ref{tab:sota}. We can see there is an improvement of $\sim$3\% upto $\sim$27\% in eight datasets.  
The Previous Best scores and their methods from the following works respectively; $\lozenge$~\cite{phandi-etal-2018-semeval}: BiLSTM for MDB-SENTCLS and Rule Based method for MDB-RELCLS,$\blacklozenge$~\cite{Queiroz2019DetectingHT}: CNN + Word Embedding method, $\bigstar$~\cite{webist/MohassebAK20}: Random Forest classifier with SMOTE algorithm, $\clubsuit$~\cite{Marchal2014PhishStormDP}: Random Forest, $\spadesuit$~\cite{aaai/SatyapanichFF20-CASIE}: pure-built BERT method for EVTDET and Noe Event Specific system for ARGROLE, $\heartsuit$~\cite{simran2019deep}: Bidirectional GRU+CNN-CRF model, and $\triangle$~\cite{Tabassum20acl}: SOFTNER (BERTOverflow).
The performance, however, drops by 2\% for SOFTNER. We believe the use of domain specific embeddings by the SOFTNER authors helped.

\smallskip
\smallskip
\noindent
\textbf{R2: To what extent Task Transfer is possible with UTS in few-shot settings?}

\begin{table}[]
\centering
\small
\resizebox{\linewidth}{!}{%
\begin{tabular}{@{}lrrrr@{}}
\toprule
     Dataset        & FS-20 & FS-50 & FS-100 & T5-FL \\ \midrule
CASIE-EVTARG (DKTU) &       65.90	&66.23	&67.64	&69.89           \\
SAL (DKTR)         &      89.31	&89.63	&89.73	&90.42           \\
SOFT-NER (DKTR)    &      78.60	&78.22	&80.45	&80.85           \\
Soft-Flaw NER (DUTR)   &      50.10	&53.16	&54.95	&76.71           \\ \bottomrule
\end{tabular}%
}
\caption{Entity Extraction (EE) Task Transfer - FS: few-shot UTS on 20, 50, 100 samples, T5-FL: T5 on full}
\label{tab:eett}
\end{table}

\noindent
Table \ref{tab:eett} shows the performance of various settings of EE task transfer. We can see for both DKTR and DKTU  settings, even with only 20 samples, UTS can achieve performance very close (within $\sim$2\%) to T5-base trained on full data. This shows even though model is not explicitly trained for a task it can perform well with very few examples if it is trained on similar training data. However, for DUTR, the model achieves ($\sim$50 F1 points) for FS-20 but falls quite short ($\sim$12 F1 points) of the T5-FL setting. This shows that task transfer becomes challenging when data nature changes.

\begin{table}[]
\centering
\small
\resizebox{\linewidth}{!}{%
\begin{tabular}{@{}lrrrr@{}}
\toprule
    Dataset         & FS-20 & FS-50 & FS-100 & T5-FL \\ \midrule
CASIE-EVTARG (DKTU) &  86.26     &  94.61     & 96.09       & 97.94           \\
SAL (DKTR)          &  1.06     &  3.46     & 85.86       &    99.44        \\
SOFT-NER (DKTR)    &  28.65     &  33.97     & 42.67       & 76.69           \\\bottomrule
\end{tabular}%
}
\caption{Entity Typing (ET) Task Transfer - FS: few-shot UTS on 20, 50, 100 samples, T5-FL: T5 on full}
\label{tab:ettt}
\end{table}

Table \ref{tab:ettt} shows the performance of various settings of ET task transfer. This task is harder for the models since the model has to generate the types not present in the text provided. Hence it can be seen that the FS-20 performance is poor as compared to T5-FL for DKTR categories. For the DKTU category, the model has the knowledge from two unrelated task of assigning role to event arguments and detecting events. Even though the role categories do not have any overlap with arguments, the model still have some understanding of the argument type from these two unrelated tasks.

For both the tasks, EE and ET, UTS's performance increase with more samples in all settings. We also notice that, T5-base performed poorly for each few-shot settings (appendix Tables \ref{tab:eet5} and \ref{tab:ett5}).

\smallskip
\smallskip
\noindent
\textbf{R3: To what extent Domain Transfer is possible with UTS in few-shot settings ?}

\begin{table}[]
\tiny
\resizebox{\linewidth}{!}{%
\begin{tabular}{@{}lrr@{}}
\toprule
           & Soft-Flaw (CLS) & Soft-Flaw (NER) \\ \midrule
FS-20 (UTS)     &        82.14 &	50.00         \\
FS-50 (UTS)     &        82.17 &	61.54                \\
FS-100 (UTS)    &        82.21 &	65.67              \\
Supervised (T5-FL) &        83.63 &	76.71               \\ \bottomrule
\end{tabular}%
}
\caption{Domain Transfer on Twitter Dataset, Supervised: Trained with T5-Base on full dataset, Fewshot(FS) with 20, 50, 100 samples.}
\label{tab:domain_transfer}
\end{table}
\noindent
Table \ref{tab:domain_transfer} shows how much domain transfer can UTS perform with twitter texts. For Soft-Flaw CLS dataset FS-20 performance is within $\sim$1.5\% F1 of T5-FL while Soft-Flaw NER dataset the model falls quite short of the T5 full dataset trained model. This, we believe, is because classification is an easier task than NER for generative models and also UTS has more supervision from classification datasets than NER.


\smallskip
\smallskip
\noindent
\textbf{R4: Is it possible to perform Zero-shot Domain Transfer with UTS ?}
We explore if UTS can be adapted to another domain for the same task or same domain for some other tasks in zero-shot setting. We find that for the classification task (CLS), UTS can predict whether a text is `malicious' or not with 82.92\% F1 which is less than 1\% short of T5-FL. This is marginally greater than FS-100 performance. A possible explanation can be that these few-shot datasets are label balanced and the model learns well for the positive labels and not so well for the negative labels.

\section{Case Studies}
We analyze the prediction output of UTS for each datasets. Here we present a few of them.

\begin{figure}[ht!]
    \centering
    \includegraphics[width=0.48\textwidth]{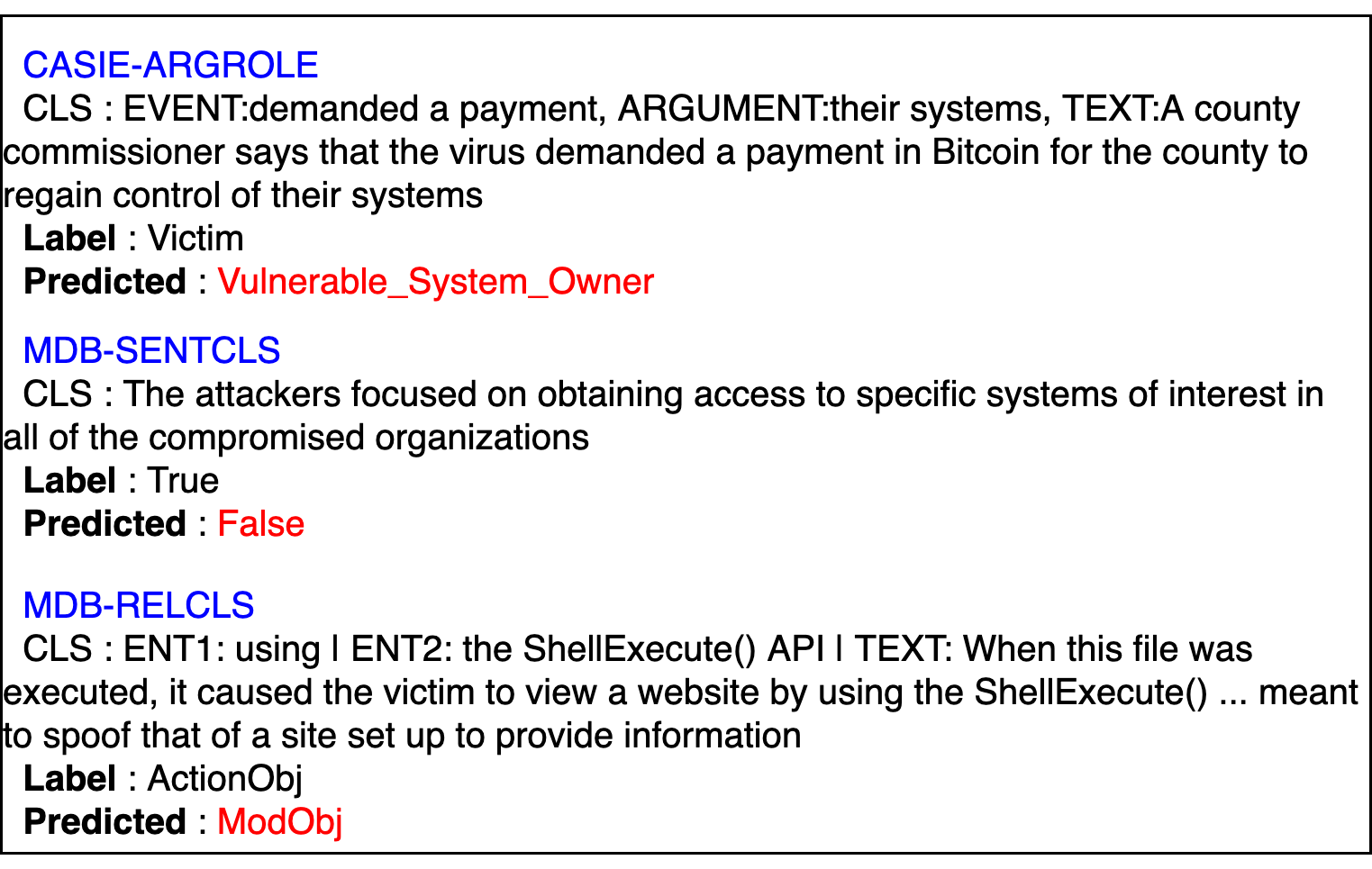}
    \caption{CLS Example Predictions: MDB and CASIE}
    \label{fig:error_cls}
\end{figure}

\noindent
\textbf{Classification:}
Figure \ref{fig:error_cls} shows one example from each classification dataset where our model fails. Here, the first example (CASIE Event Role) is interesting since the classification decision is quite close. The model understood that `their system' argument is a vulnerable system but failed to understand that it is not the owner. In the MDB-RELCLS example the relation between the two entities `using' and `the ShellExecute() API' should be Action Object rather than Modifier Object.

\begin{figure}[ht!]
    \centering
    \includegraphics[width=0.48\textwidth]{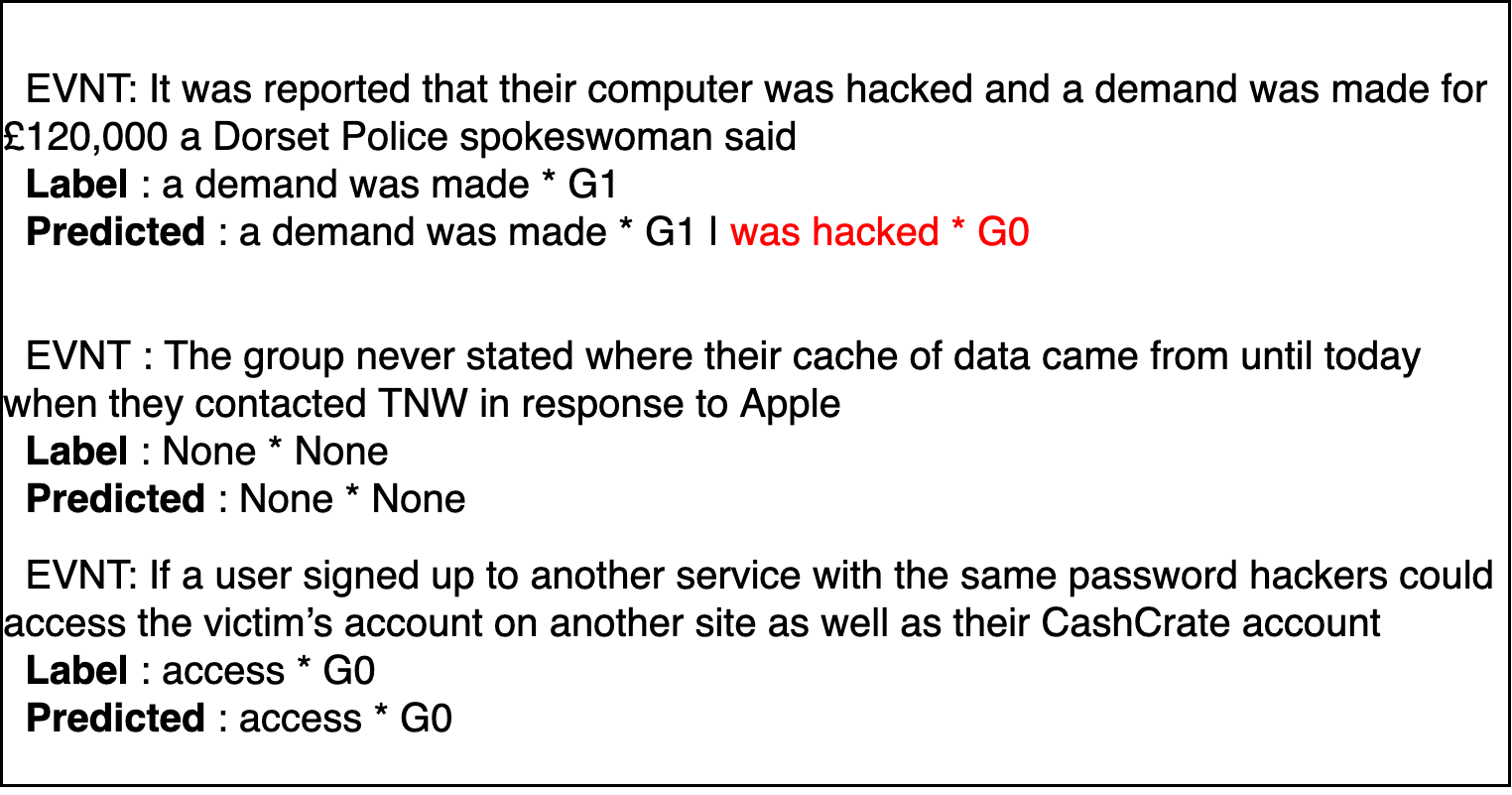}
    \caption{Event Detection: CASIE}
    \label{fig:error_evnt}
\end{figure}

\noindent
\textbf{Event Detection:}
Figure~\ref{fig:error_evnt} shows some examples of successful and incorrect prediction cases of the CASIE Event Detection dataset.
In the first example, two events are detected out of which one is correct but the other is difficult to understand by the model.
Here ``was hacked * Databreach'' is not in the gold label since ransomware attacks do not always link to databreach. In addition, CASIE only has five event types (Databreach, Phishing, Ransom, Vulnerability (discover), and Vulnerability (patch).These five types do not cover the whole cybersecurity events such as Malware, Virus, Trojan, and Spyware. Thus, we suspect that our model detected a potential event phrase ``was hacked'' and assigned one of the five event types even if there is no suitable type in the candidates.

\begin{figure}[ht!]
    \centering
    \includegraphics[width=0.48\textwidth]{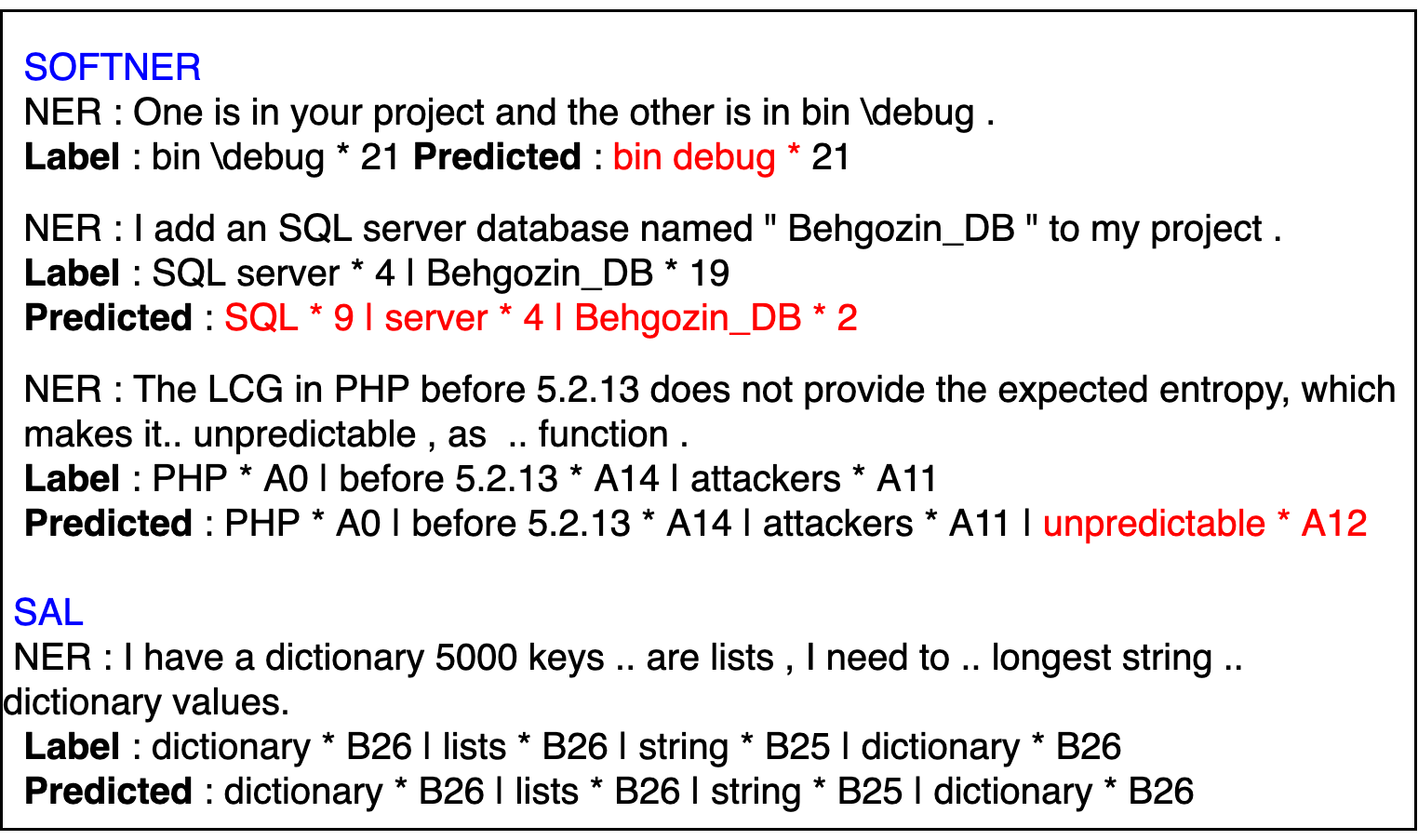}
    \caption{Named Entity Recognition}
    \label{fig:error_ner}
\end{figure}

\smallskip
\noindent
\textbf{Named Entity Recognition:}
Figure~\ref{fig:error_ner} shows four examples of NER task. The first three examples are from SOFTNER dataset and the last one from SAL dataset. The first example shows that our model predicted words and entity type correctly. However, the special character ``$\backslash$'' is missing from the prediction. Since we use exact-match metrics for evaluation this categories of incorrect prediction penalizes the models. We also find similar examples where our models could not generate full entities with other characters like `$\lbrace$' and `$\rbrace$'. The second example has two issues; the first one is split `SQL server' into `SQL' and `server' and assigned different entity to `SQL' part, the second one is that the word/phrase predicted correctly, however, the entity type is incorrect. 
We also find cases to last example, where UTS has correctly predicted more than three entities and their types.

\begin{figure}[ht!]
    \centering
    \includegraphics[width=0.48\textwidth]{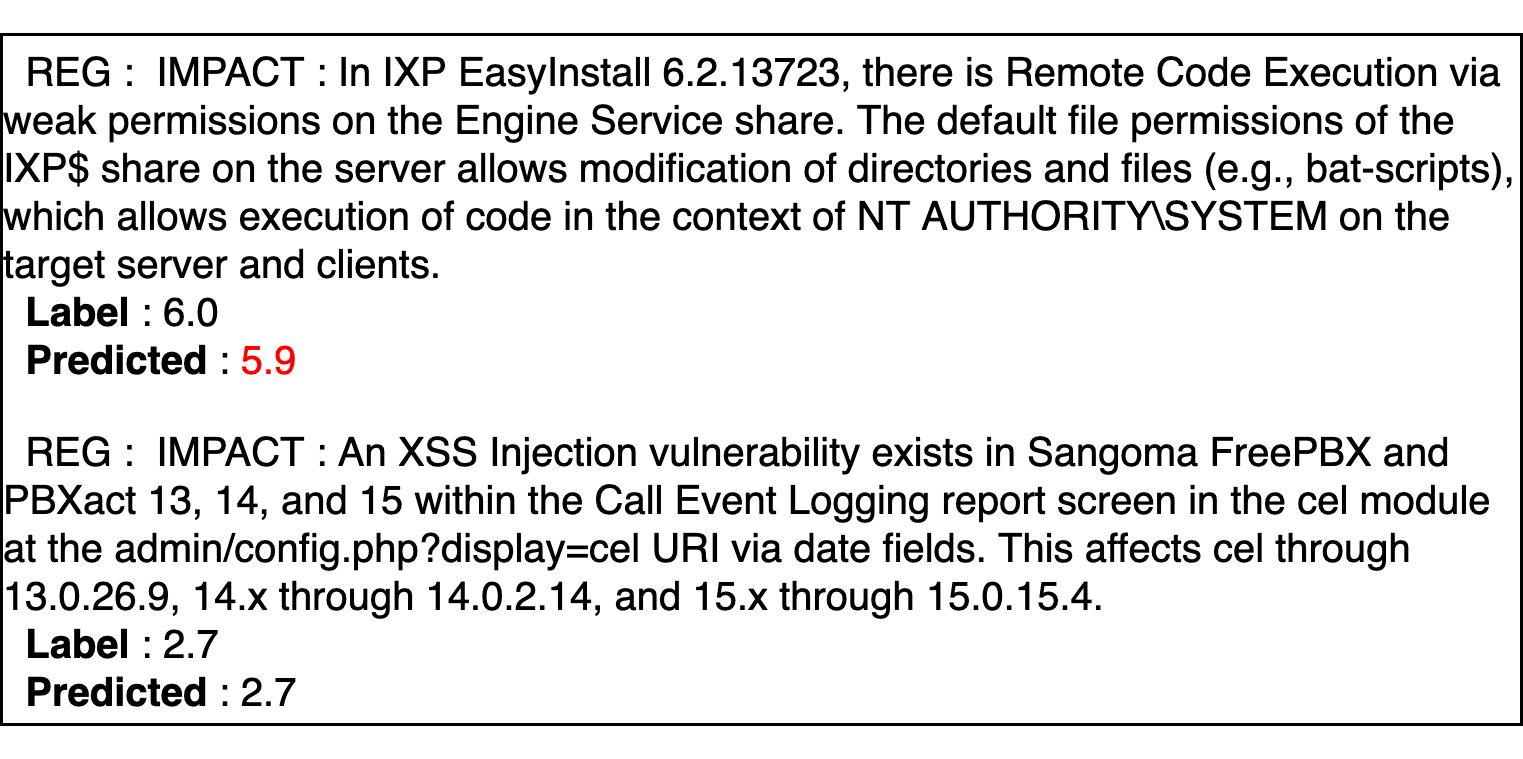}
    \caption{Regression : Impact Score}
    \label{fig:error_reg_impact}
\end{figure}

\smallskip
\noindent
\textbf{Regression:}
Figure~\ref{fig:error_reg_impact} shows successful and unsuccessful predictions of the regression task. While the original Impact Scores are calculated based on several features in the vulnerability, our model predicts these scores based on only from the textual description of the vulnerability (CVE descriptions). The second example show that UTS missed to predict the actual impact score by a close margin.

\section{Related Work}
\textbf{Multitask Learning in Diverse domains:}
\noindent
In natural language domain, DecaNLP~\cite{mccann2018natural} introduced the approach of converting multiple task into single QA format to train and evaluate ten tasks. With the gradual introduction of stronger generative NLP models like GPT, T5 and BART, the text-to-text unified models gained prominence. The multi-task approach has been shown to perform well in various domains like SciFive~\cite{phan2021scifive} in the biomedical domain, CodeT5~\cite{wang2021codet5} in the source code domain,  LEGAL-BERT~\cite{chalkidis2020legal} in legal domain and FinBERT~\cite{ijcai/0001HH0Z20} in financial service domain.
Using ``teacher forcing'' for all tasks for training with a maximum likelihood objective, SciFive enables multitask learning.
CodeT5 is a unified pre-trained encoder-decoder Transformer model and it can handle various tasks across various directions between program languages and natural languages.

\noindent
\textbf{Task-Based Unified Models:}
\noindent
Apart from these, there are individual task based unified models like 
InstructionNER which expands the existing methods for sentence-level tasks to a instruction-based generative framework for low-resource named entity recognition~\cite{corr/abs-2203-03903}. 
In biomedical domain, KGNER ~\cite{health/BanerjeePDB21} formulated the NER task as a multi-answer knowledge guided question-answer task and experimented with 18 datasets.
UnifiedNER~\cite{yan2021unified} works on unifying span-based, nested and discontinuous NER tasks. UnifiedQA~\cite{khashabi2020unifiedqa} showed that an unified training of QA tasks help in improvement of other QA tasks. Similar results are shown in common-sense reasoning tasks by Unicorn~\cite{lourie2021unicorn}.

\noindent
\textbf{NLP Approaches on Cybersecurity:}
\noindent
NLP approaches have been applied in cybersecurity domain on various nature of texts involving function-calls, software binaries and network traffics~\cite{uss/ChuaSSL17,sp/ZhangYYTLKAZ21,sigsoft/PeiGBCYWUYRJ21}. 
There are approaches that apply to specific cybersecurity tasks like lexical analysis of domain name~\cite{cybersecpods/KidmoseSP18}, a syntactic analysis (parsing)~\cite{bigdataconf/PereraHBDW18}, keyword extraction for phishing classification~\cite{isi/LHuillierHWR10}, NER based automated system to diagnose cybersecurity situations in Internet of Thing (IoT) networks~\cite{sensors/GeorgescuIZ19}.
There are many systems using social media, blog posts and discussion forums for analyzing and extracting CTI(Cyber Threat Intelligence) information as well as measuring the risk of vulnerability exploitation~\cite{compsec/ZhaoYLSHL20,zenebe2019cyber,deliu2018collecting,deliu2017extracting,uss/SabottkeSD15,www/PortnoffADKBMLP17,almukaynizi2017predicting,almukaynizi2019patch,almukaynizi2020logic,corr/abs-2102-07869}.
In addition, there have been works in extracting flow structure from unstructured software vulnerability analysis discussions in public forums~\cite{acl/PalKBMWB21}. However, our focus here is to unify varied nature of texts and  introduce an unified approach in this domain.



\section{Conclusion and Future Work}
In this work, we introduce a multi-nature, multi-task approach, UTS, in the cybersecurity domain. We experiment  with T5-base, a transformer-based generative model and show that the unified approach shows significant improvements on two datasets when compared with individual training. Also, it improves over most of the previous best performances. We show that task transfer is possible when the UTS model is trained with fewer samples of training data. This indicates that UTS can be adapted to new tasks and only few training samples are necessary. We believe this will reduce the annotation costs for new tasks. We also show that UTS is robust to the new nature of texts and also requires few samples to adapt. In future, apart from the four fundamental NLP tasks,  we would like to add more tasks such as multi-label classification or relation extraction. We believe the approach and the benchmarks we establish can be used as a baseline for future studies in the cybersecurity domain. The experiments we perform with this UTS approach, are limited to using either  natural language texts or text with embedded source code constructs as input. There is a potential to include system calls or binary codes in these unified cybersecurity models.



\section*{Limitations}
We presented a multitask model trained jointly with limited data in the cybersecurity domain.
There are some limitations to our work. First, we work with multiple nature of texts aggregating which is a challenge. In this research, our focus is on unifying mostly variations of textual nature along with some embedded software code constructs. We do not include other nature of cybersecurity texts like source code, binaries, decompiled code or network traffic. Second, we have not included datasets from other languages (NER datasets in Russian texts) which also pose challenges to train in a multi-task setting and might require multi-lingual approaches. Third, for the few-shot experiments we randomly chose some examples from each category to make the dataset label-balanced. Selecting the few-shot examples might lead to minor variations in performance. Fourth, few of the older datasets have no explicit test set mentioned in their paper. For adapting them to our approach, we randomly chose 20\% as a test set leading to difference in comparison. Hence, we include individual T5 trained baseline as a comparison.

\section*{Ethics Statement}
All our experiments are performed with well-known publicly released model transformer-based model T5. We work on 13 different datasets published in notable peer-reviewed works. We do not create, collect or process these datasets in any way such that they can be considered unethical. Our trained model checkpoints will be able to perform common natural language processing tasks similar to a general natural language processing domain.


\bibliography{anthology,custom}
\bibliographystyle{acl_natbib}

\appendix

\section{Appendix}
\label{sec:appendix}

\smallskip
\smallskip
\noindent



\subsection{Other case studies:}

\textbf{NER Task Error Analysis:}

SAL Dataset:\\

\fbox{\begin{minipage}{17em}
\textbf{Text: }\\
The embedded HTTP server in multiple Lexmark laser and inkjet printers and MarkNet devices, including X94x, W840, T656, N4000, E462, C935dn, 25xxN, and other models, allows remote attackers to cause a denial of service (operating system halt) via a malformed HTTP Authorization header.\\

\textbf{Gold: }\\
Lexmark * N | X94x * F | W840 * F | T656 * F | N4000 * F | E462 * F | C935dn * F | 25xxN * F | allows * L | remote attackers * L | denial of service * L | Authorization * L\\

\textbf{Predicted: }\\
Lexmark * N \textcolor{red}{| inkjet * O | MarkNet * A | X94x * O | W840 * O | T656 * O | N4000 * O | E462 * O | C935dn * O | 25xxN * O |} allows * L | remote attackers * L | denial of service * L | Authorization * L
\end{minipage}}\\
\\

\fbox{\begin{minipage}{17em}
\textbf{Text: }\\
Certain patch-installation scripts in Oracle Solaris allow local users to append data to arbitrary files via a symlink attack on the /tmp/CLEANUP temporary file , related to use of Update Manager.\\

\textbf{Gold: }\\
Solaris * I | local users * L | arbitrary files * L | symlink attack * L\\

\textbf{Predicted: }\\
\textcolor{red}{Oracle * N | Solaris * A |} local users * L | arbitrary files * L | symlink attack * L
\end{minipage}}
\\

\textbf{ED Task Error Analysis:}

CASIE Event Detection dataset:\\

\fbox{\begin{minipage}{17em}
\textbf{Text: }\\
It was reported that their computer was hacked and a demand was made for £120,000 a Dorset Police spokeswoman said\\

\textbf{Gold: }\\
a demand was made * Ransom\\

\textbf{Predicted: }\\
a demand was made * Ransom \textcolor{red}{| was hacked * Databreach}
\end{minipage}}\\
\\

\fbox{\begin{minipage}{17em}
\textbf{Text: }\\
EVNT : The group never stated where their cache of data came from until today when they contacted TNW in response to Apple\

\textbf{Gold: }\\
None * None\\

\textbf{Predicted: }\\
None * None
\end{minipage}}\\

\fbox{\begin{minipage}{17em}
\textbf{Text: }\\
Launched in 2016 the No More Ransom scheme brings law enforcement and private industry together in the fight against cybercrime and has helped thousands of ransomware victims retrieve their encrypted files without lining the pockets of crooks\\

\textbf{Gold: }\\
Ransom * Ransom\\

\textbf{Predicted: }\\
\textcolor{red}{The No More} Ransom \textcolor{red}{scheme} * Ransom
\end{minipage}}\\

CASIE Event Arguments dataset:\\

\fbox{\begin{minipage}{17em}
\textbf{Text: }\\
The attack disabled servers early Tuesday morning, and city officials say it was contained by 5:30 PM Wednesday.\\

\textbf{Gold: }\\
servers * System | early Tuesday morning * Time | 5:30 PM Wednesday * Time\\

\textbf{Predicted: }\\
\textcolor{red}{disabled servers} early Tuesday morning * \textcolor{red}{Capabilities}
\end{minipage}}\\
\\

\fbox{\begin{minipage}{17em}
\textbf{Text: }\\
In some cases, a generic password is required, although security researchers have discovered that in many cases, FTP servers can be accessed without a password.\\

\textbf{Gold: }\\
FTP servers * System | can be accessed without a password * Capabilities | security researchers * Person\\

\textbf{Predicted: }\\
security researchers * Person | \textcolor{red}{FTP servers} can be accessed without a password * Capabilities
\end{minipage}}\\

\textbf{Classification Task Error Analysis
}

MalwareTextDB v2: \textit{Sentence Classification}\\

\fbox{\begin{minipage}{17em}
\textbf{Text: }\\
The Skelky ( from skeleton key ) tool is deployed when an attacker gains access to a victims network ; the attackers may also utilize other tools and elements in their attack.\\

\textbf{Gold: }\\
False\\

\textbf{Predicted: }\\
\textcolor{red}{True}
\end{minipage}}\\
\\

\fbox{\begin{minipage}{17em}
\textbf{Text: }\\
The attackers focused on obtaining access to specific systems of interest in all of the compromised organizations.\\

\textbf{Gold: }\\
True\\

\textbf{Predicted: }\\
\textcolor{red}{False}
\end{minipage}}\\
\\

MalwareTextDB v2: \textit{Relation Classification}\\

\fbox{\begin{minipage}{17em}
\textbf{Text: }\\
a tool | allows  | <doc>.\\

\textbf{Gold: }\\
SubjAction\\

\textbf{Predicted: }\\
\textcolor{red}{CoRefer}
\end{minipage}}\\
\\

MalwareTextDB v2: \textit{Attribute Classification}\\

\fbox{\begin{minipage}{17em}
\textbf{Text: }\\
executing | After the C\&C reply, Moose continues with infection, executing commands on the victim device.\\

\textbf{Gold: }\\
Capability TacticalObjectives StrategicObjectives
\\

\textbf{Predicted: }\\
Capability \textcolor{red}{ActionName}  StrategicObjectives TacticalObjectives
\end{minipage}}\\
\\

\fbox{\begin{minipage}{17em}
\textbf{Text: }\\
obtaining |     The attackers focused on obtaining access to specific systems of interest in all of the compromised.\\

\textbf{Gold: }\\
Capability StrategicObjectives
\\

\textbf{Predicted: }\\
Capability
\end{minipage}}\\
\\

\fbox{\begin{minipage}{17em}
\textbf{Text: }\\
allows |          On January 12, 2015, Dell Secureworks blogged about a tool (Trojan.Skelky) that allows attackers to...\\

\textbf{Gold: }\\
Capability StrategicObjectives\\

\textbf{Predicted: }\\
\textcolor{red}{ActionName}
\end{minipage}}\\
\\

\textbf{Regression Task Error Analysis
}\\

NVD CVE metrics: \textit{Impact score}\\

\fbox{\begin{minipage}{17em}
\textbf{Text: }\\
An issue was discovered in B\&R Industrial Automation APROL before R4.2 V7.08. Some web scripts in the web interface allowed injection and execution of arbitrary unintended commands on the web server, a different vulnerability than CVE-2019-16364.\\

\textbf{Gold: }\\
5.9\\

\textbf{Predicted: }\\
\textcolor{red}{3.6}
\end{minipage}}\\
\\

NVD CVE metrics: \textit{Exploitability score}\\

\fbox{\begin{minipage}{17em}
\textbf{Text: }\\
Libspiro through 20190731 has a stack-based buffer overflow in the spiro\_to\_bpath0() function in spiro.c.\\

\textbf{Gold: }\\
2.2\\

\textbf{Predicted: }\\
\textcolor{red}{2.8}
\end{minipage}}\\
\\

\begin{table}[]
\centering
\small
\resizebox{\linewidth}{!}{%
\begin{tabular}{@{}lrrrr@{}}
\toprule
     Dataset        & FS-20 & FS-50 & FS-100 & T5-FL \\ \midrule
CASIE-EVTARG (DKTU) &       0.00	& 0.00	& 11.44	&69.89           \\
SAL (DKTR)         &      0.00	& 0.04	& 71.34	&90.42           \\
SOFT-NER (DKTR)    &      46.2	& 20.85	& 38.35	&80.85           \\
Soft-Flaw-NER (DUTR)   &      0.00	& 0.00	& 53.73	&76.71           \\ \bottomrule
\end{tabular}%
}
\caption{Entity Extraction (EE) Task Transfer - FS: few-shot T5 base model on 20, 50, 100 samples, T5-FL: T5 on full}
\label{tab:eet5}
\end{table}

\begin{table}[]
\centering
\small
\resizebox{\linewidth}{!}{%
\begin{tabular}{@{}lrrrr@{}}
\toprule
    Dataset         & FS-20 & FS-50 & FS-100 & T5-FL \\ \midrule
CASIE-EVTARG (DKTU) &  23.65     &  33.50     & 71.52       & 97.94           \\
SAL (DKTR)          &  0.20     &  0.07     & 72.89       &    99.44        \\
SOFT-NER (DKTR)    &  21.64     &  16.96     & 22.53       & 76.69           \\\bottomrule
\end{tabular}%
}
\caption{Entity Typing (ET) Task Transfer - FS: few-shot T5 base model on 20, 50, 100 samples, T5-FL: T5 on full}
\label{tab:ett5}
\end{table}

\subsection{Case Study}
\begin{figure}[ht!]
    \centering
    \includegraphics[width=0.48\textwidth]{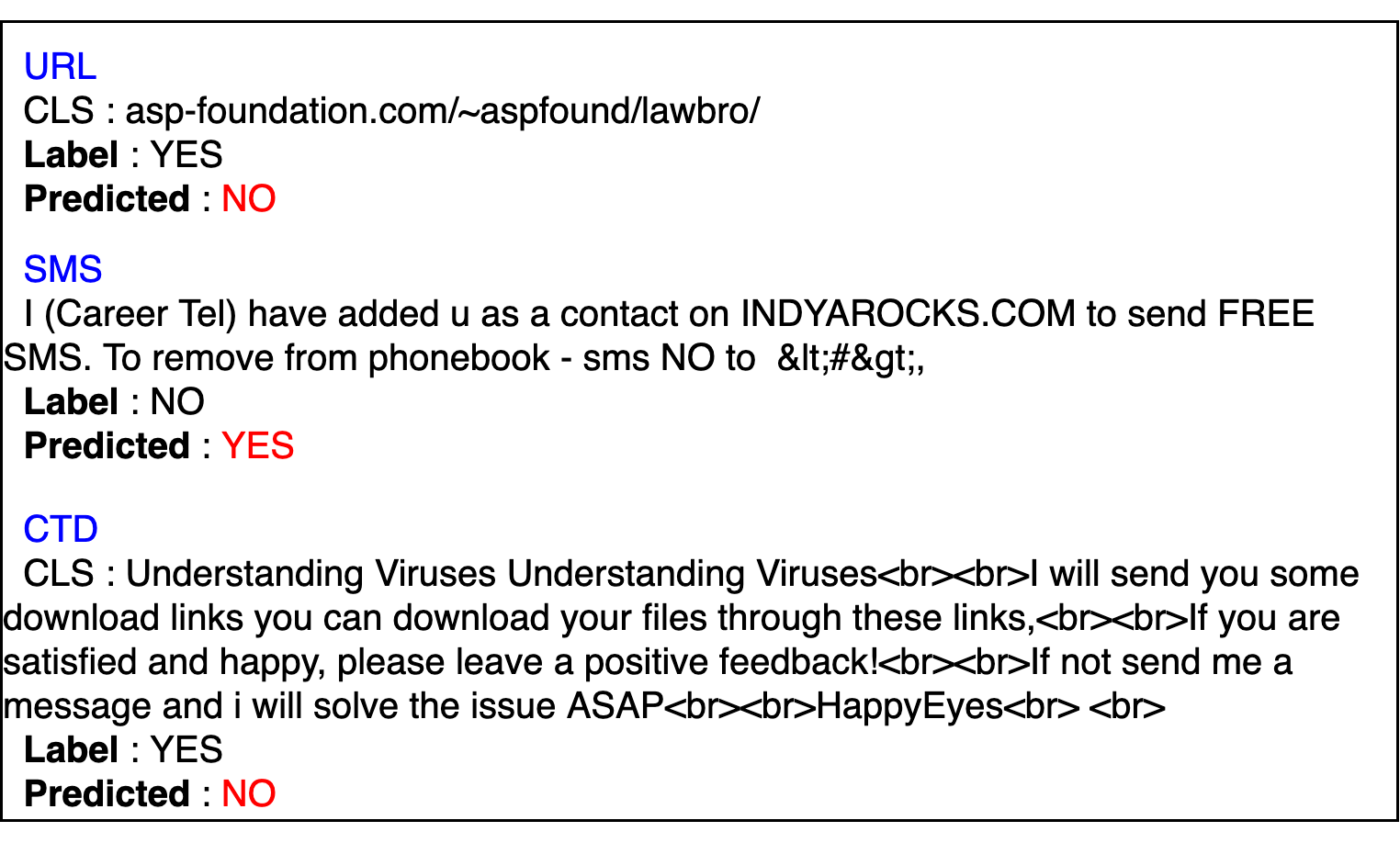}
    \caption{Classification Example Predictions : URL, SMS, CTD}
    \label{fig:error_cls2}
\end{figure}

\noindent
\textbf{Classification:}
Figure~\ref{fig:error_cls2} shows more cases of failed classification in URL, SMS and CTD datasets. 

\noindent
\textbf{Event Detection:}
The second example in Figure~\ref{fig:error_evnt} shows the empty case that the given text does not have any event. Our model predicted ``None * None'' correctly. In other words, our model is trained not to predict if there is no event in the given text. In addition, our model predicted exactly as the third example in Figure~\ref{fig:error_evnt} shows.


\begin{figure}[ht!]
    \centering
    \includegraphics[width=0.48\textwidth]{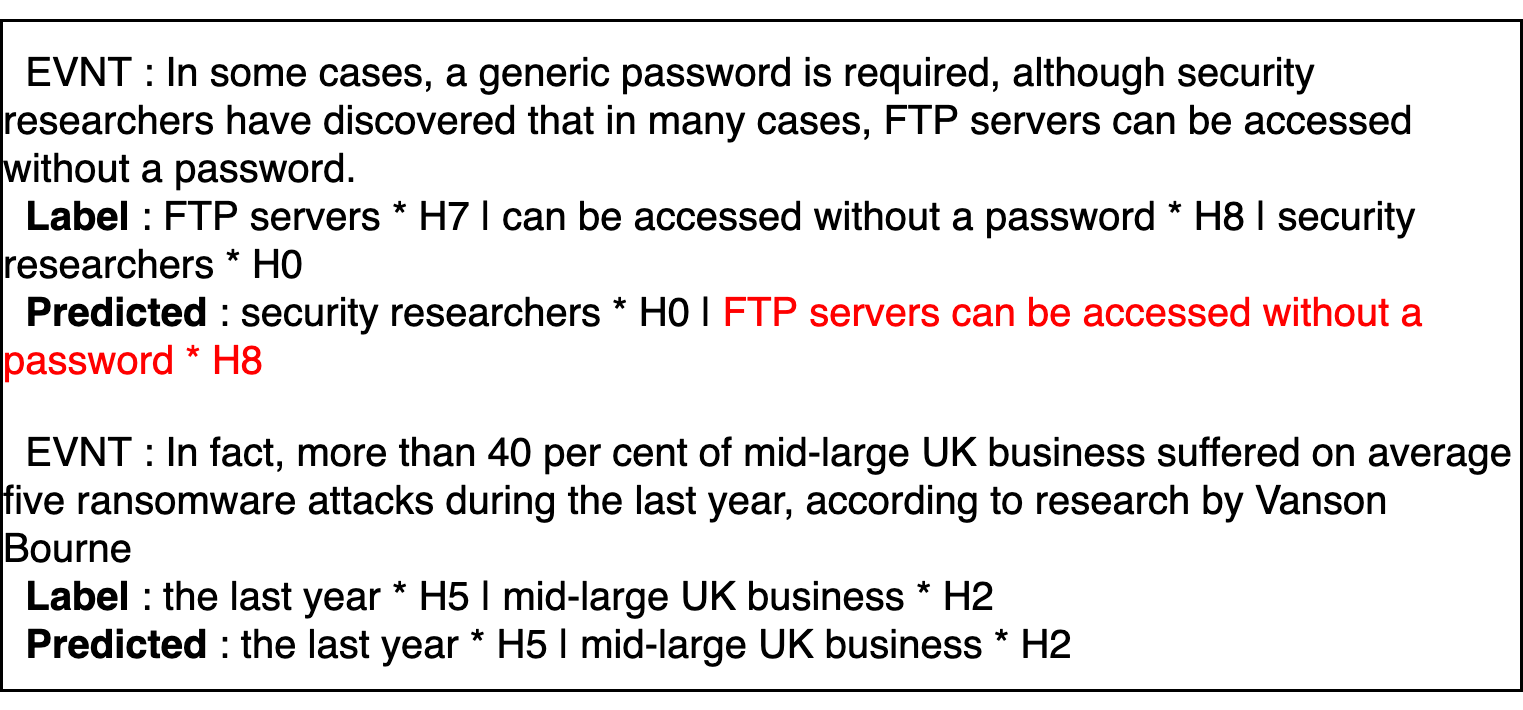}
    \caption{Event Argument Detection : CASIE}
    \label{fig:error_evntarg}
\end{figure}

Figure~\ref{fig:error_evntarg} shows some examples of success and failure cases of the CASIE Event Argument Detection dataset.
Our model predicted ``security researchers * Person | FTP servers can be accessed without a password * Capabilities'' from the given text ``In some cases, a generic password is required, although security researchers have discovered that in many cases, FTP servers can be accessed without a password.'' The gold label is ``FTP servers * System | can be accessed without a password * Capabilities | security researchers * Person'', and the predicted phrase ``FTP servers can be accessed without a password'' combined ``System'' part and ``Capabilities'' part.

On the other hand, our model can predict multiple arguments from the given text as the second example of Figure~\ref{fig:error_evntarg} shows.

\noindent
\subsection{Libraries Used:}
For building and training $UTS$, we use publicly available packages : PyTorch \cite{NEURIPS2019_9015} 1.9.1, HuggingFace Transformers \cite{wolf-etal-2020-transformers} 4.15.0, HuggingFace Datasets \cite{lhoest-etal-2021-datasets} 1.16.1, seqeval \cite{seqeval} 1.2.2, sklearn \cite{scikit-learn} 1.0 and pandas 1.3.4.



\end{document}